\documentclass[a4paper,UKenglish,cleveref, autoref, thm-restate]{lipics-v2021}

\usepackage{tikz}
\usepackage{caption}
\usepackage{booktabs}
\usepackage{cleveref}
\usepackage{amsmath}

\newcommand{\setactivities}{\mathcal{A}}
\newcommand{\setteams}{\mathcal{W}}

\newcommand{\naturalnumber}{\mathbb{N}}

\hideLIPIcs  

\title{Trustworthy and Explainable Decision-Making for Workforce allocation}

\author{Guillaume {Povéda}}{Airbus SAS, FR}{guillaume.poveda@airbus.com}{https://orcid.org/0000-0001-9175-3240}{}
\author{Ryma Boumazouza}{Airbus SAS, FR}{ryma.boumazouza@airbus.com}{https://orcid.org/0000-0002-3940-8578}{}{}
\author{Andreas Strahl}{Airbus Aerostructures, GmbH}{andreas.strahl@airbus.com}{}{}{}
\author{Mark Hall}{Airbus Operations Ltd, UK}{mark.hall@airbus.com}{}{}{}
\author{Santiago Quintana-Amate}{Airbus Operations Ltd, UK}{santiago.quintana-amate@airbus.com}{https://orcid.org/0000-0002-9308-6717}{}{}
\author{Nahum Alvarez}{Airbus SAS, FR}{nahum.alvarez@airbus.com}{https://orcid.org/0000-0003-1717-2506}{}{}
\author{Ignace Bleukx}{DTAI, KU Leuven, Leuven, Belgium}{ignace.bleukx@kuleuven.be}{https://orcid.org/0000-0001-7810-8351}{}{}
\author{Dimos Tsouros}{DTAI, KU Leuven, Leuven, Belgium}{dimos.tsouros@kuleuven.be}{https://orcid.org/0000-0002-3040-0959}{}{}
\author{Hélène Verhaeghe}{DTAI, KU Leuven, Leuven, Belgium}{helene.verhaeghe@kuleuven.be}{https://orcid.org/0000-0003-0233-4656}{}{}
\author{Tias Guns}{DTAI, KU Leuven, Leuven, Belgium}{tias.guns@kuleuven.be}{https://orcid.org/0000-0002-2156-2155}{}{}

\authorrunning{G. Povéda et al.} 

\Copyright{Guillaume Povéda et al.} 



\category{} 

\relatedversion{} 

\acknowledgements{This work has been partially funded by the European Union’s Horizon Europe Research and Innovation program under the
grant agreement TUPLES No 101070149}

\nolinenumbers 

\EventEditors{John Q. Open and Joan R. Access}
\EventNoEds{2}
\EventLongTitle{42nd Conference on Very Important Topics (CVIT 2016)}
\EventShortTitle{CVIT 2016}
\EventAcronym{CVIT}
\EventYear{2016}
\EventDate{December 24--27, 2016}
\EventLocation{Little Whinging, United Kingdom}
\EventLogo{}
\SeriesVolume{42}
\ArticleNo{23}

\begin{document}

\maketitle

\begin{abstract}
In industrial contexts, effective workforce allocation is crucial for operational efficiency. This paper presents an ongoing project focused on developing a decision-making tool designed for workforce allocation, emphasizing the explainability to enhance its trustworthiness. Our objective is to create a system that not only optimises the allocation of teams to scheduled tasks but also provides clear, understandable explanations for its decisions, particularly in cases where the problem is infeasible. By incorporating human-in-the-loop mechanisms, the tool aims to enhance user trust and facilitate interactive conflict resolution. We implemented our approach on a prototype tool/digital demonstrator intended to be evaluated on a real industrial scenario both in terms of performance and user acceptability. 
\end{abstract}

\section{Introduction}
In industrial contexts, effective workforce allocation is a cornerstone of operational efficiency, directly impacting productivity, cost management, and overall organizational performance. The complex nature of workforce allocation involves balancing numerous constraints, such as employee availability, skill levels, regulatory requirements, and task priorities. As industries increasingly rely on automated decision-making tools to manage these complexities, the need for trustworthiness and explainability in these systems becomes paramount.

This paper introduces an ongoing project dedicated to the development of a decision-making tool tailored to workforce allocation. The core objective of this tool is to not only optimise the allocation of teams to scheduled tasks but also to ensure that the decision-making process is transparent and understandable to users. 
Current industrial workforce allocation often functions as a black box, primarily due to the complexity and opacity of the underlying processes. 
This lack of transparency hinders the general understandability of the solution and is detrimental to the development and deployment of automatic solutions using AI tools. 
Our work aims to address this problem by improving transparency and explainability of workforce allocation systems. 
Complicating workforce allocation processes, are the need for real-time adaptation of the workforce under disruptions.
The necessary knowledge to manage these disruptions is often implicit, `hidden' in the planners' heads, making it difficult for AI-generated solutions to gain acceptance unless they can clearly explain their rationale. 
Our approach not only seeks to enhance the transparency of workforce allocation but also aims to ensure that AI solutions can effectively communicate their decision-making processes, thereby increasing trust and acceptance among human planners.

Another significant challenge in workforce allocation is the occurrence of infeasible situations, where the constraints cannot be satisfied simultaneously. Traditional systems may simply fail or produce sub-optimal solutions without providing clear explanations, leading to user frustration and mistrust. To overcome this, our tool incorporates human-in-the-loop mechanisms, enabling users to interact with the system to understand and resolve infeasibilities. These explainability features are designed to enhance user trust and facilitate effective conflict resolution, making the decision-making process more collaborative and reliable.

In summary, this paper presents an integrated approach to workforce allocation, emphasizing the importance of trustworthiness and explainability. By integrating interactive features and human-in-the-loop mechanisms, we aim to create a decision-making tool that is not only effective but also transparent and user-friendly, paving the way for more reliable and collaborative industrial operations.

Looking ahead, future plans include evaluating the tool's effectiveness. This evaluation will focus on assessing the tool's impact on operational efficiency, user understandability and acceptance, and its ability to handle real-world workforce allocation scenarios.


\subsection{Overview of workforce allocation challenges}
In the industrial landscape, efficient workforce allocation or task scheduling is a critical component of operational success. 
We consider here the operational problem of assigning teams of workers, to a set of \emph{already scheduled} tasks, in a manner that optimises workers utilization and meets various operational constraints. 
Furthermore, workers have different availability slots; in real scenarios, uncertainty (represented by accidents, illnesses or simply time delays in other tasks) may further modify this pre-established availability. 

To address this, we have already developed a decision-making tool relying on constraint programming (CP)~\cite{rossi2006handbook}, a powerful paradigm well-suited for solving complex allocation problems. While we will describe this tool in detail in the next section, it is important to note that even with a highly performant solver, eXplainable AI (XAI) is essential to ensure the trustworthiness and acceptance of AI solutions in workforce allocation. 

Despite the technical robustness of CP solvers, their adoption in industrial settings is often hindered by a perceived lack of transparency and lack of user interaction capabilities. 
Decision-makers and end-users frequently struggle to understand the rationale behind the solver’s outputs, particularly when the problem is infeasible. 
This can lead to mistrust and underusage of the technology, ultimately diminishing its potential benefits. 
Also, the actual modelling of the problem may be challenging as the modelling experts are often not the final users of the decision-making tool. 

To overcome these challenges, our project focuses on integrating explainability and trustworthiness into the CP-based decision-making tool. 
By providing clear, comprehensible explanations for the solver’s decisions and highlighting reasons for infeasibilities, we aim to build greater user trust and facilitate more effective human-computer collaboration. 
Interactive features are also being developed to allow users to engage with the tool, explore alternative solutions, and iteratively restore feasibility when conflicts arise.

This paper outlines our ongoing efforts to create an explainable and trustworthy workforce allocation tool. We demonstrate the implementation of interactive conflict resolution mechanisms and discuss our plans for evaluating these features. 

\subsection{Explainability in Constraint Programming}
Explainability in AI has evolved significantly over time, driven by the need to make AI systems more transparent, trustworthy, and user-friendly. The authors in \cite{cyras2020machine} broadly categorized the questions that explanations in AI aim to answer into three classes: \textit{What and Why} (What made/Why did the system reach this outcome?), \textit{Why not and What if} (Why did the system not reach a different outcome? What if different information were used?), and \textit{How} (How can I modify the system to obtain a more desirable outcome with the existing information?). 
This categorization helps understand the progression and focus of explainability efforts in various AI methodologies, including machine reasoning (MR) and machine learning (ML). 
Different methodologies have addressed these explainability questions (see \cite{gonul2006effects,johnson1993explanation,kulesza2015principles,lacave2002review,lim2010toolkit,miller2019explanation,mohseni2021multidisciplinary,molnar2020interpretable,moulin2002explanation,preece2018asking,samek2019explainable,sokol2020explainability}).

The remainder of this section focuses on the specific application of XAI techniques within the domain of constraint programming, especially in workforce allocation and scheduling problems.

Constraint Programming is a powerful method at the intersection of AI and OR, for solving combinatorial problems. 
CP involves specifying constraints that need to be satisfied and finding solutions that meet these constraints. 
Explainability is crucial in CP, particularly for workforce allocation and scheduling, where decision-makers need to understand the rationale behind the allocation decisions. 
Different existing methods are used to enhance explainability in CP and can be categorized as: 
\begin{itemize}
    \item Explanation of Constraints: Making the constraints and their roles in the decision-making process clear to users. 
    \item Solution Traceability: Allowing users to trace back the steps and decisions made by the CP solver to understand how a particular solution was reached (e.g., \cite{bleukx2023simplifying}).
    \item Conflict Explanation: Identifying and explaining conflicts or infeasibilities when no solution can be found, which is particularly important for iterative problem-solving and debugging (e.g., \cite{liffiton2016fast,marques2010minimal}).
\end{itemize}

 A significant focus within explainable constraint solving is on the latter and is about explaining why a set of constraints is unsatisfiable. Many of these methods \cite{boumazouza2021asteryx,ignatiev2015smallest,junker2001quickxplain,lauffer2019human,leo2017debugging,liffiton2016fast,liffiton2008algorithms,marques2010minimal} aim to identify a minimal unsatisfiable subset (MUS) - an irreducible subset of constraints which causes the model to be unsatisfiable.

 \begin{definition}[Minimal Unsatisfiable Subset \cite{liffiton2008algorithms}]
 Given an unsatisfiable set of constraints $C$, a subset $U \subseteq C$ is a Minimal Unsatisfiable Subset if and only if $U$ is unsatisfiable and every strict subset $U' \subsetneq U$ is satisfiable
 \end{definition}
 
 Such explanations are interesting because they pinpoint the exact constraints responsible for the inconsistency, allowing users to focus their efforts on resolving specific issues. 
 Recently, research has also been directed towards advising users on how to restore feasibility \cite{gupta2022finding,senthooran2021human}, notably by identifying the minimal correction subset (MCS) \cite{boumazouza2020symbolic}. 

\begin{definition}[Minimal Correction Subset \cite{liffiton2008algorithms}]
Given an unsatisfiable set of constraints $C$, a subset $M \subseteq C$ is a Minimal Correction Subset if an only if $C \setminus M$ is satisfiable, and for every strict subset $M' \subsetneq M$, $C \setminus M'$ is unsatisfiable.
\end{definition}
 
 An MCS is particularly useful because it identifies an irreducible set of constraints that, when modified or relaxed, can restore the feasibility of the entire system. 
 By focusing on such a minimal set, users can implement the least disruptive changes necessary to resolve conflicts, which helps maintain the integrity of the original constraint problem as much as possible. 
 However, there remains a shortage of tools that effectively explain why a problem is inconsistent. 




\section{Problem definition}
The problem consists of assigning teams of workers to tasks in a large-scale industrial setting, involving several hundreds of daily activities. 
We will consider the set of tasks to accomplish as already scheduled in time, each of them needs to be allocated to a team of workers. Any given team of workers can't be allocated to two activities at the same time neither do 2 tasks in a row when there is some geographical constraint such transportation time that makes it impossible. Each team has its own calendar of availability or set of skills that can restrict the set of activities it can be allocated to. In this section, we will introduce the needed notations and formulate the base constraint model implemented to solve it:
\subsection{Notations}
\begin{enumerate}
    \item $\setactivities$ the set of activities to accomplish
    \item $\setteams$ the set of worker teams available 
    \item $\forall a\in \setactivities$, $start_a\in \naturalnumber , end_a \in \naturalnumber$, the start and end time of the activity $a$
    \item $\forall a\in \setactivities$, $comp_a \in 2^{\setteams}$ stores the subset of worker teams compatible with the activity $a$. Similarly we can define binary indicator $comp\_binary_{a,w}\in \{0,1\}, \forall a \in \setactivities, \forall w\in \setteams$ storing the same information.
    \item $\mathcal{S}$ is a list of activity pair $(a_i, a_j)$ that should be allocated to the same team.
\end{enumerate}

\subsection{Constraint model}
In this section, we detail the CP formulation implemented for the problem. A Boolean formulation showed the best performance using the different solvers we tested in our backend application (like Ortools CP-SAT \cite{ortools}, Exact \cite{Exact}, and Gurobi \cite{gurobi}). 
\subsubsection*{Variables}
\begin{enumerate}
    \item Let $\forall a\in \setactivities, w\in \setteams, alloc_{a,w}\in \{0,1\}$ be the allocation variable. A value of $1$ will correspond to given worker team $w$ being allocated to the activity $a$.
    \item Let $\forall w \in \setteams, used_w \in \{ 0,1 \}$, be the Boolean variable indicating if a given team $w$ is allocated to any of activities $a\in \setactivities$
\end{enumerate}

\subsubsection*{Constraints}
\begin{enumerate}
    \item \label{c-task-allocated} Each task in allocated : $\forall a\in \setactivities, \sum_{w\in \setteams}{alloc_{a,w}}=1$
    \item \label{c-pair-overlap} Non-Overlapping constraint : 
    
    $\forall a\in \setactivities$ we denote $neigh(a)=\{a' \in \setactivities \ s.t \ (end_{a'}>start_{a}) \land end_a \geq start_{a'} \}$ the set of overlapping activities of activity $a$, then $\forall w \in \setteams, a'\in neigh(a), alloc_{a,w}+alloc_{a',w}\leq 1$
    \item \label{compatibility-team} Compatibility constraint : $\forall a\in \setactivities, w \in \setteams, \neg comp\_binary_{a,w} \rightarrow \neg alloc_{a,w}$, 
    \item \label{same-allocation} Same allocation constraint : 

    $\forall (a_1, a_2) \in \mathcal{S}, \forall w \in \setteams, alloc_{a_1,w}=alloc_{a_2,w}$
    
    \item \label{used-team} Used team constraint : $\forall a\in \setactivities, w \in \setteams, alloc_{a,w} \rightarrow used_w $
    \item \label{redundant-constraint} Aiming at speeding up solver we introduce two main additional kinds of constraint, one redundant for the overlapping constraint, and one adding symmetry breaking : 
    \begin{enumerate}
        \item \label{clique-constraint}Clique constraints : 

        $\forall a\in \setactivities$, let $overlapstart(a)=\{a'\in \setactivities, start_{a'}\leq start_{a} < end_{a'}\}$ the set of task also executed at time $start_a$ (including $a$), then this set constitutes a clique of overlapping tasks. We add the following constraint : 

        $\forall a\in \setactivities, \forall w \in \setteams, \sum_{a'\in overlapstart(a)}{alloc_{a',w}}\leq 1$

        \item \label{symmetry-constraint} Symmetry breaking: Some teams $\in \setteams$ can execute the same set of tasks for the given time horizon.
        Hence, they are equivalent and tasks can be assigned to any of those teams without changing the validity of the allocation.
        Clearly, this means equivalent teams are \emph{symmetric} and we add lexleader symmetry breaking constraints imposing an ordering of the teams~\cite{devriendt2012static, walsh2006general}.
        Several formulations are possible, but from limited testing, we found adding the ordering on the $\mathit{used}$ variables seemed most promising. It's worth noticing that this constraint will not impact solution quality, only when the objective itself treats the teams as equivalent.
        
    \end{enumerate}
\end{enumerate}

\subsubsection*{Objective functions}
The main objective of interest here will be the number of different teams used, therefore we aim at minimizing $\sum_{w\in \setteams}{used_w}$.
Several other objectives are under study, notably adding fairness objectives, and ensuring a balanced workload among the used teams. The inclusion of those objective functions has currently only been studied in the pure optimisation and performance side and not on the explainable, therefore they will not be considered in the remaining of the paper.

\begin{figure}[h]
    \centering
    \includegraphics[width=0.95\textwidth]{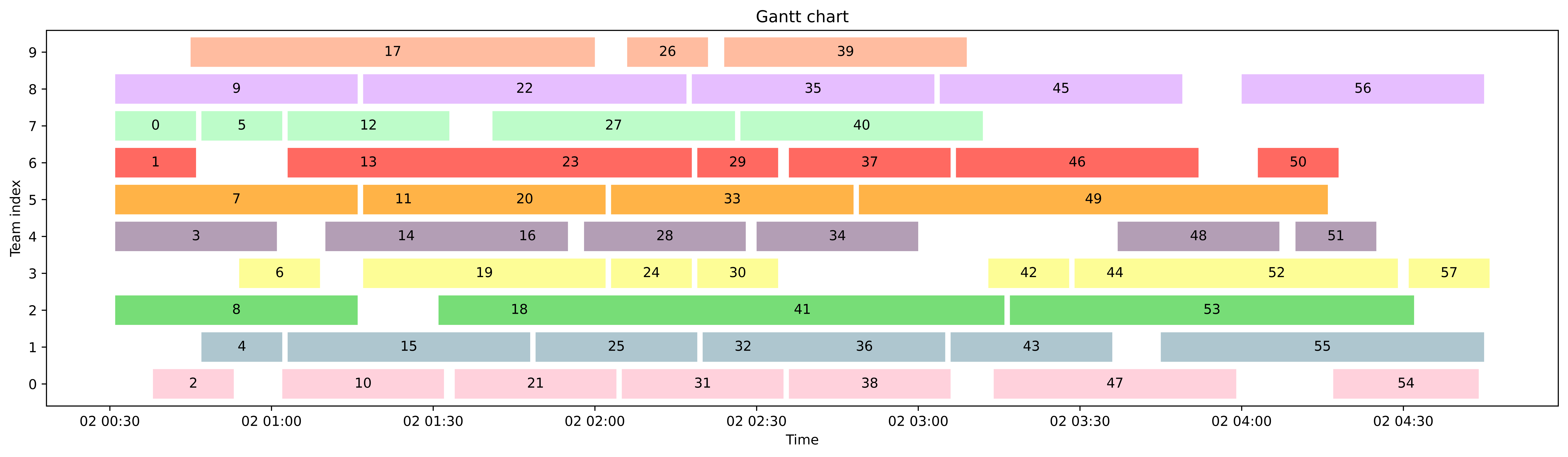}
    \caption{Example of Gantt chart built to visualise a solution to the workforce allocation problem}
    \label{gantt-ex}
\end{figure}

\paragraph*{Example of solution}
We can plot a Gantt chart to visualise the solution, as shown in Figure \ref{gantt-ex}.  Each row of the chart represents the schedule for a specific team of workers $w\in \setteams$. Due to the non-overlapping constraint (defined in constraint nb. \ref{c-pair-overlap}), a feasible solution ensures that there are no overlapping activities within each row of the Gantt chart. 
\vspace{2cm}
\section{Explainable Decision-making tool for workforce
allocation}
The development of a decision-making tool for workforce allocation is driven by the need to enhance operational efficiency, but such a tool introduces new trustworthiness requirements in order to get user acceptance. The following figure \ref{workflow} outlines the primary workflow of the tool.

\begin{figure}[h]
\centering
\includegraphics[width=0.8\textwidth]{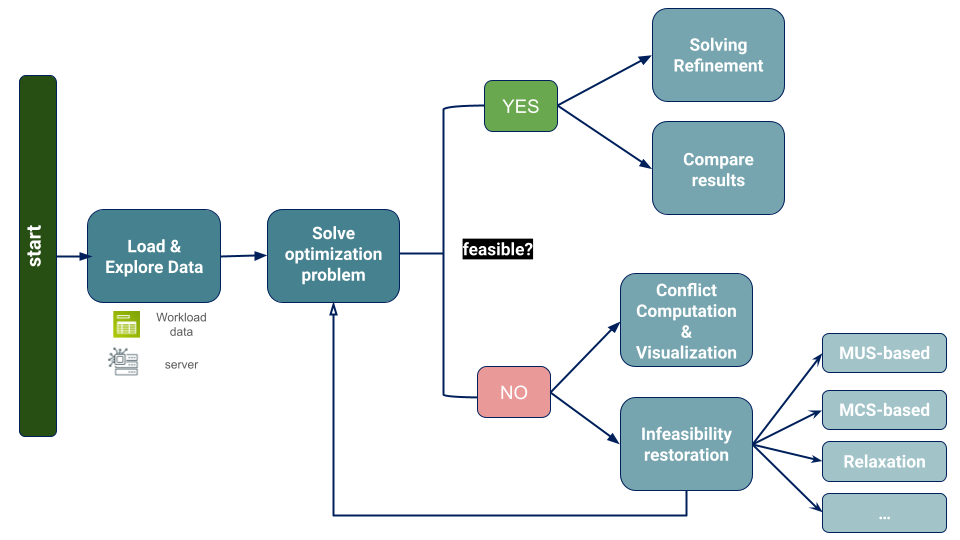}
\caption{Workflow of the Decision-Making Tool}
\label{workflow}
\end{figure}

Our tool integrates explainability components addressing two major needs: conflict computation and visualisation, and interactive infeasibility restoration. 
The explainability features of our tool are tailored to scenarios where the workforce allocation problem is infeasible, where the constraints cannot be satisfied simultaneously (e.g., when there are insufficient resources available to allocate all tasks). By addressing these infeasibility cases, the tool aims to provide insights into its decision-making process. 

\subsection{Conflicts Computation and Visualisation}

In complex allocation or scheduling scenarios, conflicts are often inevitable due to various reasons: overlapping tasks, resource constraints, and varying team availabilities. Our tool computes and visualises these conflicts, allowing users to see where and why the allocation process encounters issues. Visual representations of conflicts enable users to quickly grasp problematic areas and understand the constraints causing these issues. This transparency builds trust in the system, as users can see the logical reasoning behind the solver's decisions. Finding the best way to visualise the conflicts depends on user preferences, and this is the subject of ongoing work.




\subsection{Interactive Infeasibility Restoration}

Confronting infeasible problems is a common challenge in real-world applications \cite{chinneck2007feasibility}. Traditional CP solvers may report infeasibility without providing guidance on resolution. 
However, our tool offers the users an interactive method to solve conflicts in the problem; upon detecting an infeasible problem, users are presented with several methods to restore feasibility:

\begin{itemize}
    \item \textbf{Resolving MUS conflicts interactively (local conflict resolution):} This method involves resolving each MUS conflict one by one in an interactive manner (by selecting a constraint in the MUS to relax). Local conflict restoration refers to the process of addressing each conflict individually within its localized context, rather than attempting to solve all conflicts simultaneously. Users are guided through the process of addressing each local conflict sequentially, enabling a step-by-step restoration of feasibility.
    
    \item \textbf{Using MCS interactively (global conflict resolution):} Instead of addressing conflicts individually, this approach computes one of the minimal correction subsets (MCS) to resolve all conflicts simultaneously on a problem-wide scale. Global conflict restoration refers to the process of identifying and correcting a minimal set of constraints that, when adjusted, will restore feasibility to the entire system. In our tool, we consider the scenario where the user can choose only a subset of the relaxations provided by a single MCS, and users may want to mix-and-match constraints relaxations from different MCSes. Hence, our tool re-computes a new MCS after a user has relaxed some constraints, making the process iterative and interactive.
    
    \item \textbf{Fine-tuning task priorities (prioritized conflict resolution):} This method involves solving and optimising a relaxed version of the problem where task allocations become optional. Each task is given a priority/weight value which is taken into account in the optimisation criteria. Users can interactively change the priority level of tasks, allowing a lot of flexibility in the way the problem feasibility is restored, e.g. which tasks are more likely to remain or be removed.
\end{itemize}

By involving users in the resolution process, our tool ensures a more transparent, interactive, and trustworthy decision-making experience.


\subsection{Implementation}

The workforce allocation model was implemented using the \texttt{CPMpy} library \cite{guns2019increasing}, a flexible and user-friendly tool for modelling constraint programming (CP) problems. \texttt{CPMpy} offers an intuitive API that closely mirrors the functionality of \texttt{numpy}, making it accessible and easy to use for those familiar with numerical computing in Python. Using this modelling library allows us to test different solver backends, including \texttt{ortools-cpsat}~\cite{ortools}, \texttt{gurobi}~\cite{gurobi}, \texttt{pysat}~\cite{imms-sat18}, or \texttt{exact}~\cite{Exact, ijcai2018p180}. 
It also includes some native utilities to compute MUSes or MCSes, which we use extensively in this research for conflict analysis and feasibility restoration. 

\begin{figure}[htb]
    \centering
    \includegraphics[width=0.9\textwidth]{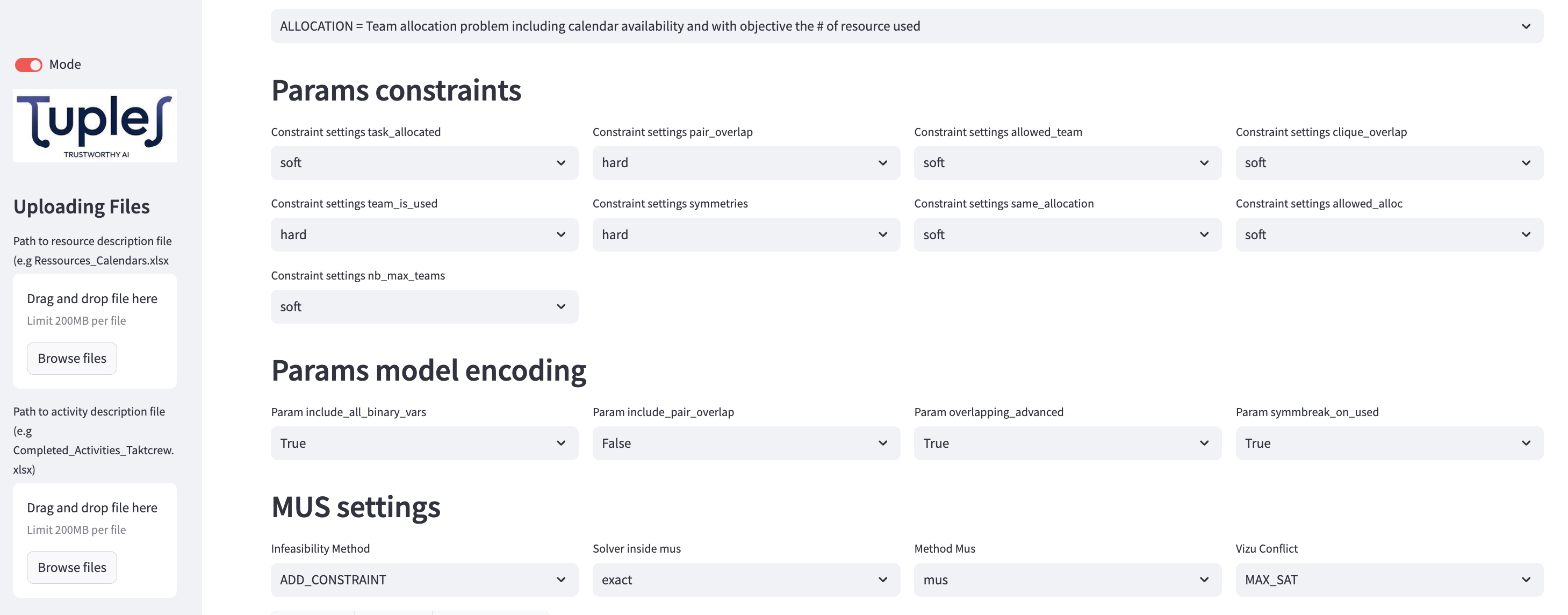}
    \caption{Configure the methods parameters tab}
    \label{parameters_view}
\end{figure}

In practice, several customization options regarding optimisation and explainability aspects are available through our configuration parameters tab within the tool, as illustrated in \Cref{parameters_view}. 

\section{Case study/Application example}
The problems to be solved in the industrial use case range from scheduling tasks over a six-hour period to creating a full day (24-hour) schedule, involving the allocation of a few dozen activities to possibly up to one thousand. The number of available resources (i.e. our $\setteams$ teams) varies over time, but typically there are about 20 ($|\setteams| \approx 20$). 


\subsection{Preliminary results}

In this section, we present our initial findings on the computational performance of the optimisation method and on the explainability components across various scenarios. For both the optimisation and explainability experiments, we generated 20 instances of the allocation problem with different lengths: 6, 8 and 24 hours. These instances were generated to reflect a real-world scenario with specific constraints and conditions derived from historical data. 
This analysis serves as a foundation for further refinement and optimisation of our approach.

\subsubsection{Optimisation results}

Despite the workforce allocation problem being NP-Hard (akin to a list colouring graph problem), preliminary empirical runs and benchmarks on historical data have demonstrated good performance. 
Our preliminary results (Table~\ref{result_merge}) consider the mean computation time to optimality (or cut to timeout) for different lengths of the instances, different CP formulations of the CP model, and different solver settings. 
The column \textbf{clique} refers to the redundant clique constraint \ref{clique-constraint} and \textbf{symmetry} to the symmetry breaking constraint on used team \ref{symmetry-constraint}.
The solver backend used is Ortools' CP-Sat solver, a state-of-the-art solver for CP problems \cite{perron2019or,perron_et_al:LIPIcs.CP.2023.3}. CP-Sat heavily relies on a portfolio approach to accomplish its search and using this feature usually will improve a lot the solving performance. To check this on our use case, we tested 2 different settings: using 1 or 6 search worker (column \textbf{\#w}). As we expected, CP-Sat is more efficient in its multi-worker settings and found optimal solution on all instances in less than 1 second in average. 
From the multi-worker settings instances, we also observed that symmetry and redundant constraints have a clear negative effect on both initialisation time of the model and on solving time.
On the contrary, in the mono-worker mode, the use of the symmetry constraints helps prove optimality for more instances and offers a computation time advantage (but still those have worse performance than their multi-worker equivalents).

These results show the efficiency of using CP-Sat with its full features activated, but it is important to note that these results do not constitute a comprehensive benchmark. 
Moreover, upon closer inspection of the results, we noticed mainly the LP-subsolver implemented in OR-tools contributed to finding a better bound during the search.
Hence, it may be interesting to evaluate the performance of LP-specific solvers too such as Gurobi.
Overall, further analysis and more extensive testing is required to validate and generalize these findings. 
Nonetheless, these results are promising and indicate the potential efficiency of CP solvers in handling complex workforce allocation problems. 

\begin{table}
\centering
\label{tab:grouped_df}
\begin{tabular}{rrrrrrrr}
\toprule
\textbf{\#w} & \textbf{len} & \textbf{clique} & \textbf{symmetry} & \textbf{t\_init(s)} & \textbf{t\_solve(s)} & \textbf{t\_total(s)} &\textbf{ optimal} \\
\midrule
1 & 6 & False & False & 0.03 & 12.21 & 12.24 & 0.43 \\
1 & 6 & False & True & 0.03 & 2.01 & 2.04 & 0.95 \\
1 & 6 & True & False & 0.06 & 10.98 & 11.04 & 0.48 \\
1 & 6 & \textbf{True} & \textbf{True} & \textbf{0.06} & \textbf{2.03} & \textbf{2.09} & \textbf{0.95} \\
1 & 8 & False & False & 0.04 & 13.13 & 13.17 & 0.45 \\
1 & 8 & False & True & 0.04 & 2.93 & 2.97 & 0.95 \\
1 & 8 & True & False & 0.08 & 12.99 & 13.06 & 0.40 \\
1 & 8 & \textbf{True} & \textbf{True} & \textbf{0.08} & \textbf{2.44} & \textbf{2.52} & \textbf{0.95} \\
1 & 24 & False & False & 0.12 & 14.64 & 14.76 & 0.30 \\
1 & 24 & False & True & 0.12 & 5.54 & 5.66 & 0.85 \\
1 & 24 & True & False & 0.33 & 14.61 & 14.94 & 0.30 \\
1 & 24 & \textbf{True} & \textbf{True} & \textbf{0.33} & \textbf{5.06} & \textbf{5.39} & \textbf{0.85} \\
\hline
6 & 6 & \textbf{False} & \textbf{False} & \textbf{0.03} & \textbf{0.06} & \textbf{0.09} & \textbf{1.00} \\
6 & 6 & False & True & 0.03 & 0.18 & 0.21 & 1.00 \\
6 & 6 & True & False & 0.06 & 0.06 & 0.12 & 1.00 \\
6 & 6 & True & True & 0.06 & 0.17 & 0.24 & 1.00 \\
6 & 8 & \textbf{False} & \textbf{False} & \textbf{0.04} & \textbf{0.08} & \textbf{0.12} & \textbf{1.00} \\
6 & 8 & False & True & 0.04 & 0.19 & 0.23 & 1.00 \\
6 & 8 & True & False & 0.08 & 0.09 & 0.17 & 1.00 \\
6 & 8 & True & True & 0.08 & 0.22 & 0.30 & 1.00 \\
6 & 24 & \textbf{False} & \textbf{False} & \textbf{0.12} & \textbf{0.29} & \textbf{0.41} & \textbf{1.00} \\
6 & 24 & False & True & 0.12 & 0.57 & 0.69 & 1.00 \\
6 & 24 & True & False & 0.34 & 0.32 & 0.66 & 1.00 \\
6 & 24 & True & True & 0.33 & 0.72 & 1.06 & 1.00 \\
\bottomrule
\end{tabular}
\caption{Mean computation time for different lengths of problem instances (\textbf{len} column), modeling parameters (\textbf{clique}, \textbf{symmetry}), and solver config on number of search worker (\textbf{\#w}))
We split the table in 2 to distinguish the multi and mono-worker settings of CPSat solver.
\textit{Remark : } Each solve call has a timeout of 30 seconds, so when optimality is not proven (like in many tests in mono-mode setting), the solve time is equal to the timeout.}
\label{result_merge}
\end{table}

\subsubsection{Explainability results}

 We run a benchmark study to compute minimal unsatisfiable subset (MUS) conflicts across various scenarios by categorizing the problem constraints into soft and hard constraints directly from our tool interface (see Figure \ref{parameters_view}). Hard constraints were necessary conditions that must be met, while soft constraints were desirable but not mandatory. This process involved determining which constraints could not be satisfied simultaneously by extracting an MUS.
 The algorithm used for finding such a MUS is based on the well-known deletion-based method~\cite{marques2010minimal}, which extracts any MUS from the problem.
 As this algorithm greatly benefits incremental solving~\cite{bleukx2023tutorial}, we used the Exact solver \cite{Exact}, a pseudo-Boolean solver which supports solving under assumptions.
 
 The study was conducted on the same instances introduced in previous sections. The evaluation focuses on the time taken to compute one explanation of infeasibility for each instance and the size of the explanation, measured in terms of the number of constraints involved in the MUS. 

 \begin{table}[h!]
    \centering
    \begin{tabular}{ccc}
    \toprule
    \textbf{Length} & \textbf{Average Time (s)} & \textbf{Average Explanation Length} \\
    \midrule
    6 & 0.60 & 10 \\
8 & 0.86 & 10 \\
24 & 1.13 & 10 \\

    \bottomrule
    \end{tabular}
    \caption{Average Calculation Time and Explanation Length by Instance Length}
    \label{result_explanation}
    \end{table}
    
The results (Table \ref{result_explanation}) indicate that as the instance length increases from 6 to 24 hours, the average calculation time for generating single explanations of infeasibility also increases (from 0.60s to 1.13s), while the average length of the explanations remains relatively consistent even for bigger instances. 
These results suggest that longer instances require more computation time, but the complexity of the explanations does not significantly increase. However, further experiments are necessary to draw definitive conclusions.

\subsection{Visualising Conflicts and Restoring Feasibility: A Practical Demonstration of our tool}
To showcase the capabilities of our explainability techniques and enable their evaluation by end users, we developed a demonstrator application using \texttt{Streamlit}\footnote{an open-source app framework for Machine Learning and Data Science projects (https://streamlit.io/)}.  
This section offers an overview of its features and functionalities.

\subsubsection{Solving the Problem}
\begin{figure}[t]
    \centering
    \includegraphics[width=0.8\textwidth]{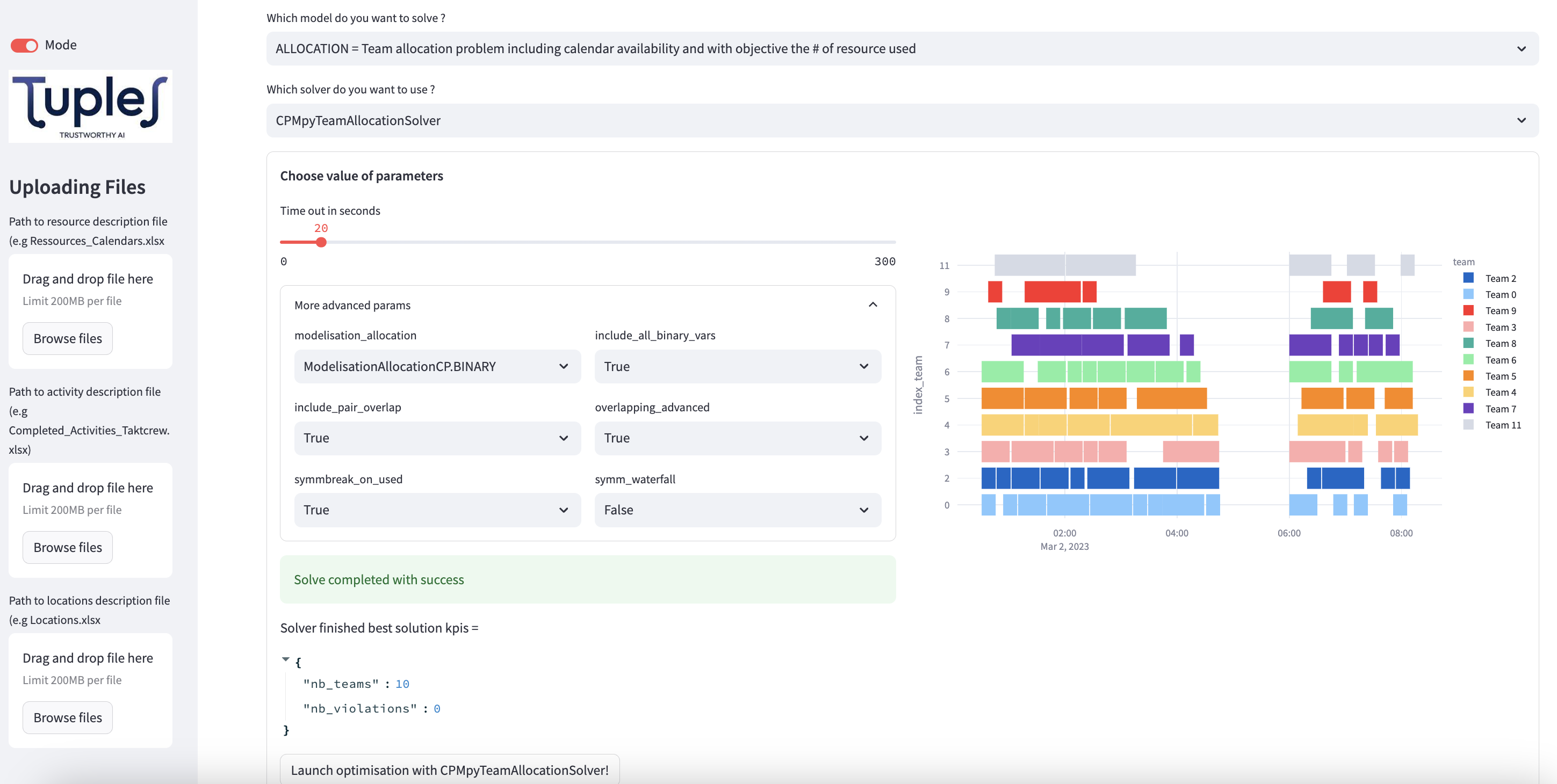}
    \caption{Solving tab of the app, showing the results after calling the solver.}
    \label{solving}
\end{figure}

The first step in our application consists of encoding and solving the allocation problem using the \texttt{CPMpy} library. Our tool allows to load data and configure various parameters, such as choosing the optimisation solver (see Figure \ref{parameters_view}). Once the problem is encoded, the solver is called to find an optimal solution. The results, including the allocation of teams to tasks, are then displayed to the user (see Figure \ref{solving}).

\subsubsection{Solution Refinement}

\begin{figure}[t]
    \centering
    \includegraphics[width=0.8\textwidth]{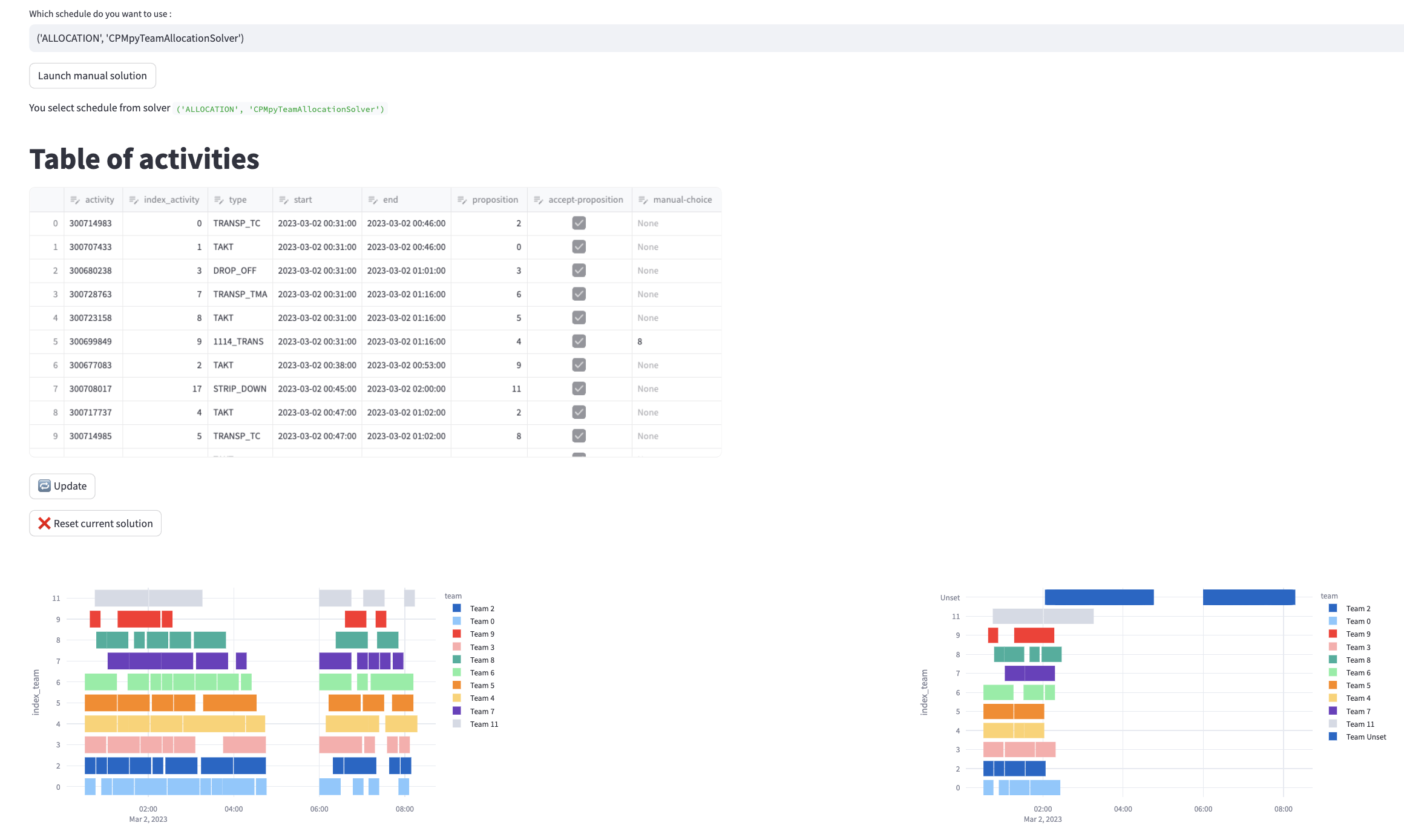}
    \caption{Interactive solving tab (Manual/Automatic)}
    \label{review}
\end{figure}

After the solver generates a solution, users have the opportunity to review it and modify it. The application allows users to propose alternative allocations overriding the solver's decisions. This interactive review process ensures that users can make adjustments based on their expertise and knowledge of the specific context (see Figure \ref{review}).

\subsubsection{Conflict Computation and Visualisation}
\begin{figure}[t]
    \centering
    \includegraphics[width=0.8\textwidth]{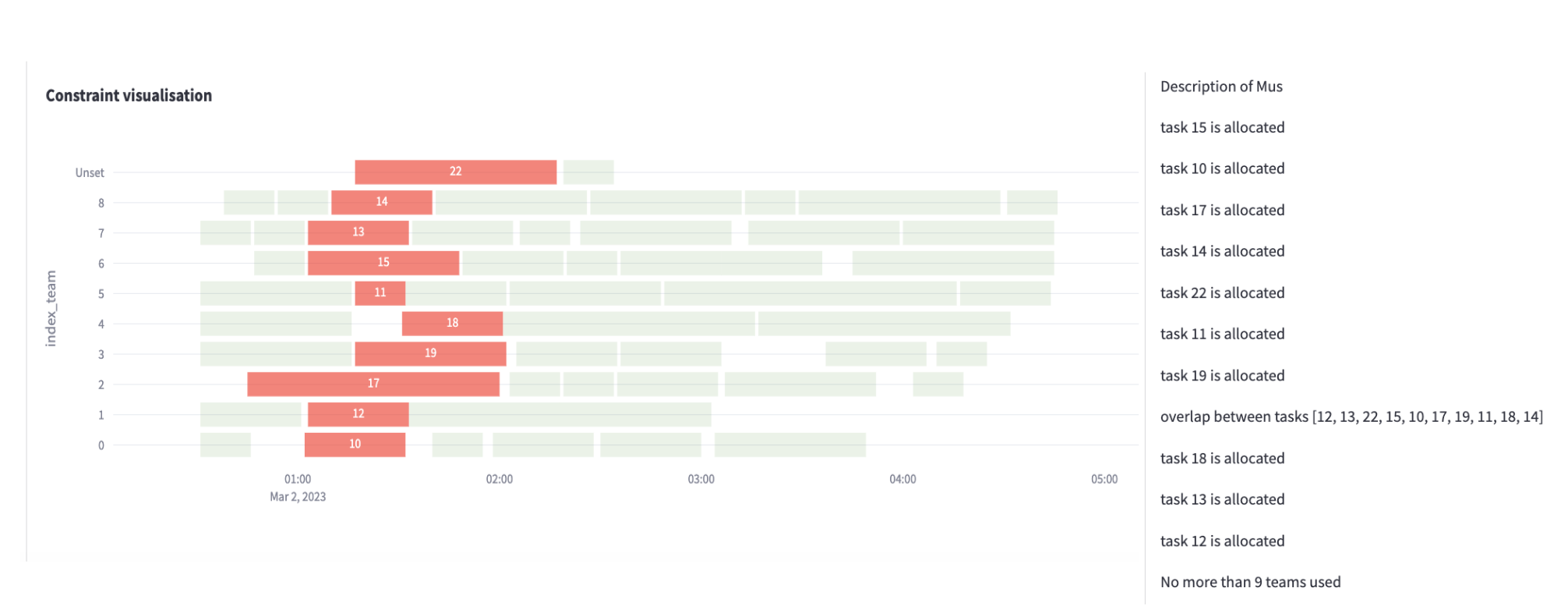}
    \caption{Conflicts visualisation solving tab, with "Unset" line at the top.}
    \label{conflicts_view}
\end{figure}

When the problem resolution is infeasible, the application computes and provides a visualisation of the conflicting constraints causing infeasibility. 
Conflicts are described with a basic text description of each of the constraints, along with a Gantt representation of the problem highlighting the activities involved in the conflict (see Figure \ref{conflicts_view}).
The displayed Gantt is built by solving an optimisation problem: it is the result of optimising the number of allocated tasks, e.g. it computes a size-maximal satisfiable subset. These tasks are then visualised, and non-allocated tasks are added to a virtual team we call "Unset", the top line of the plot. This method allows to have a visual representation even when dealing with infeasible problems where no solution (nor visualisation thereof) exists as is.

\subsubsection{Feasibility Restoration}

If the solver encounters an infeasible problem, our application offers several methods for restoring feasibility. These methods are designed to be interactive by involving the user in the resolution process.

\subsubsection*{Local Conflict Resolution}

\begin{figure}[h]
    \centering
    \includegraphics[width=0.95\textwidth]{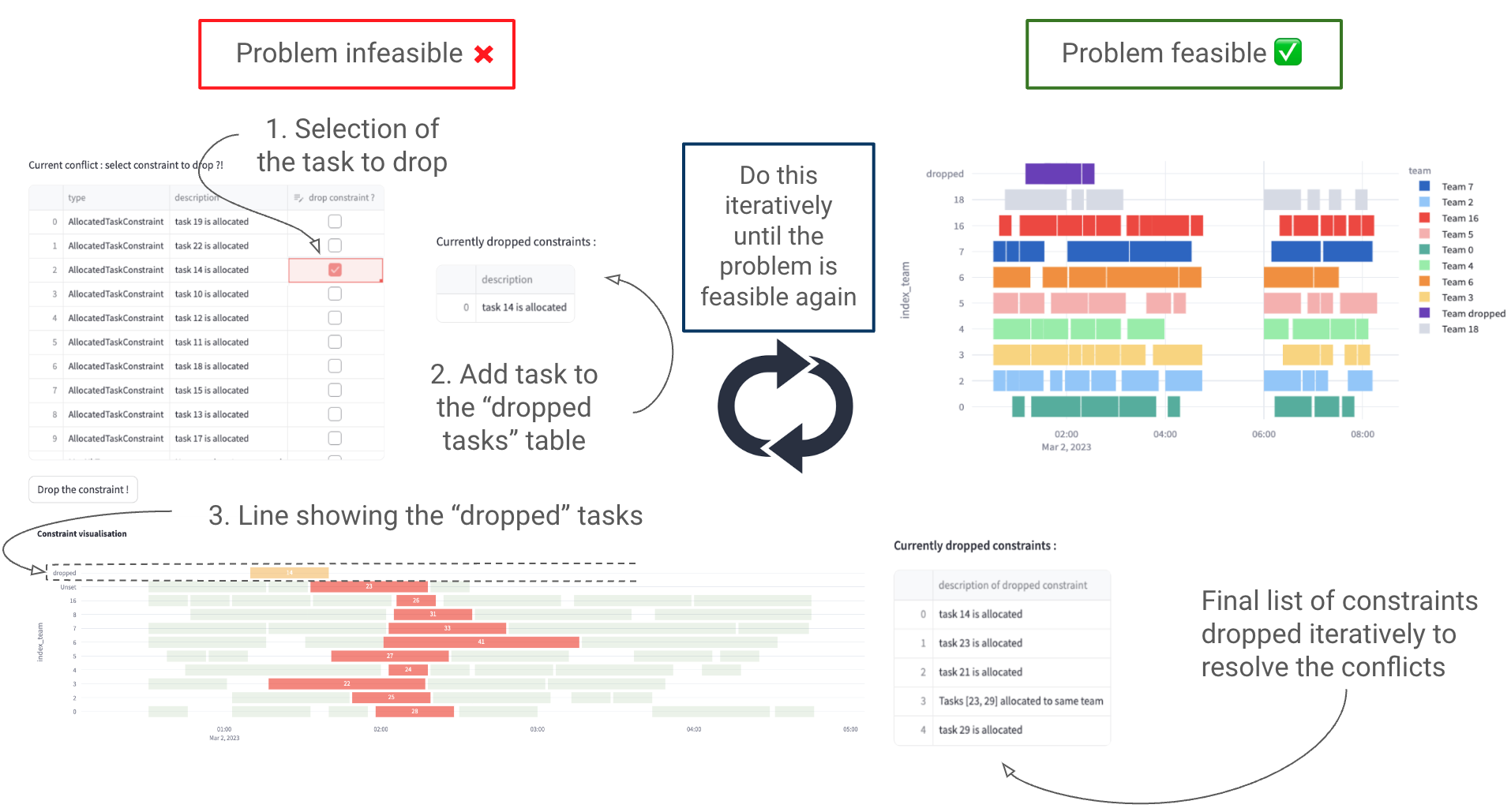}
    \caption{Process of conflict resolution with MUS }
    \label{conflicts_mus1}
\end{figure}

One approach to restoring feasibility is by resolving conflicts one by one interactively. The application identifies a minimum unsatisfiable subset (MUS) of constraints and guides users through the process of addressing each conflict individually. This local resolution method allows users to make targeted adjustments. The process is illustrated in Figure \ref{conflicts_mus1}. In our preliminary experiment, similarly to the scenario depicted, few iterations were required to restore feasibility, and we surmise that this observation remains true for real scenarios.

\subsubsection*{Use of Minimum Correction Subset}

\begin{figure}[h]
    \centering
    \includegraphics[width=0.8\textwidth]{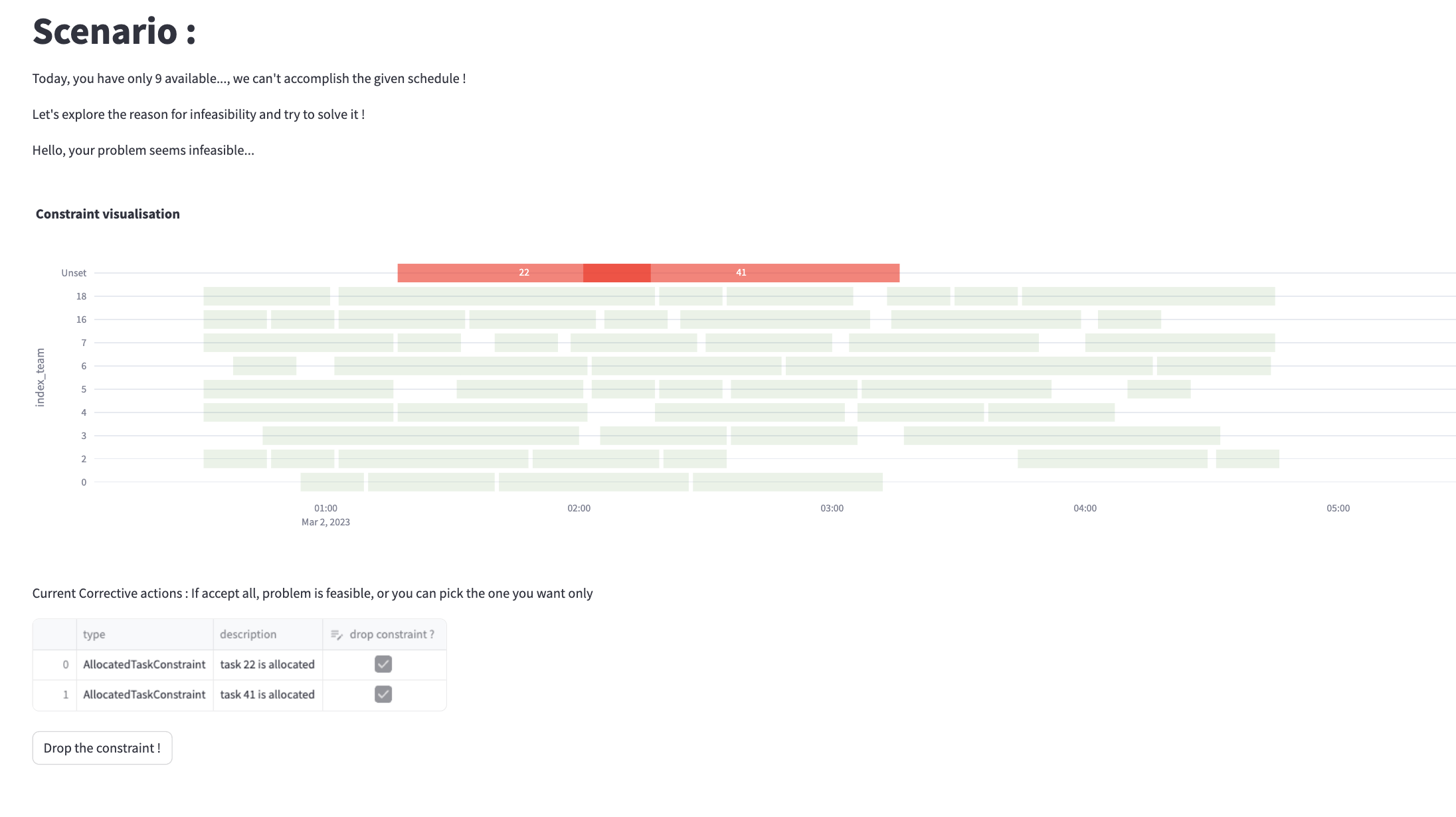}
    \includegraphics[width=0.8\textwidth]{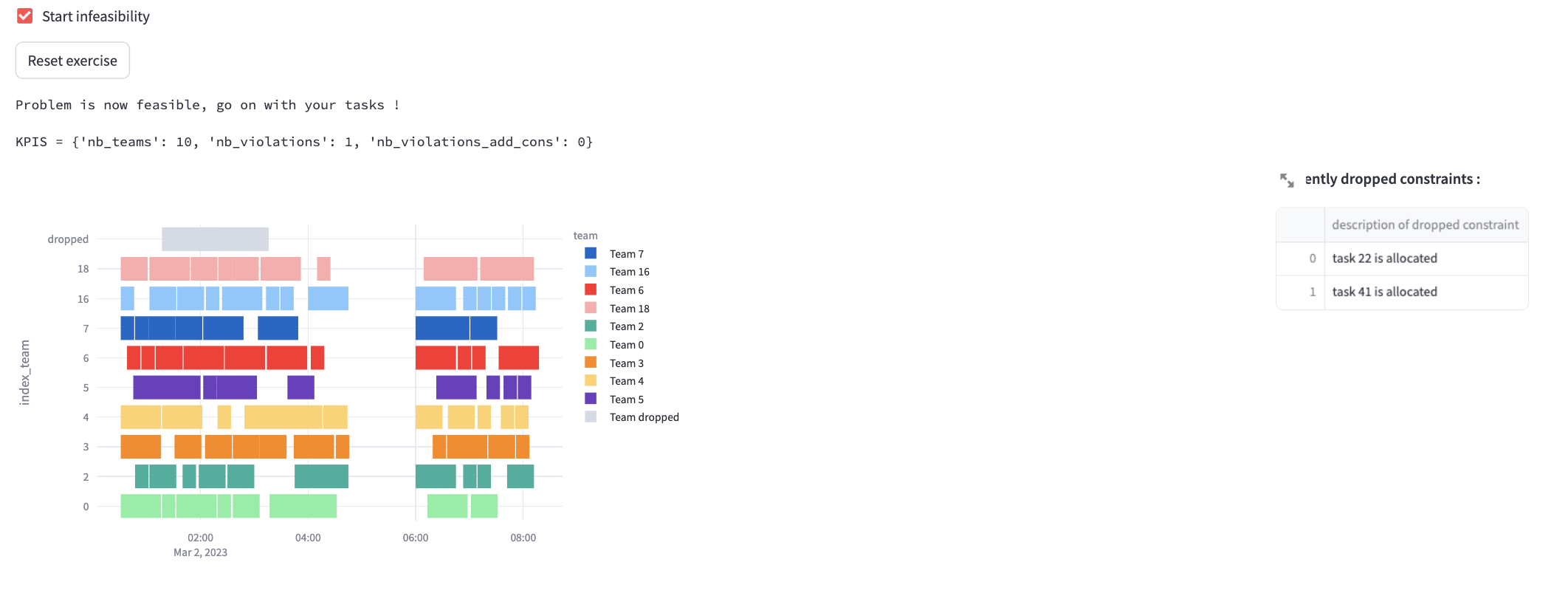}
    \caption{Conflict resolution with MCS}
    \label{conflicts_mcs}
\end{figure}

Alternatively, users can employ a minimal correction subset (MCS) to resolve conflicts globally as shown in Figure \ref{conflicts_mcs}. The application identifies a minimal set of constraints that need to be corrected to restore full feasibility. In the interactive setup, we consider that users can accept to remove only a subset of the constraints proposed by the tool (which would not completely restore the feasibility). We envisage providing multiple MCSs in the future if none of the corrective actions fits user preference.

\subsubsection*{Fine-tuning task priorities}

\begin{figure}[h]
    \centering
    \includegraphics[width=0.9\textwidth]{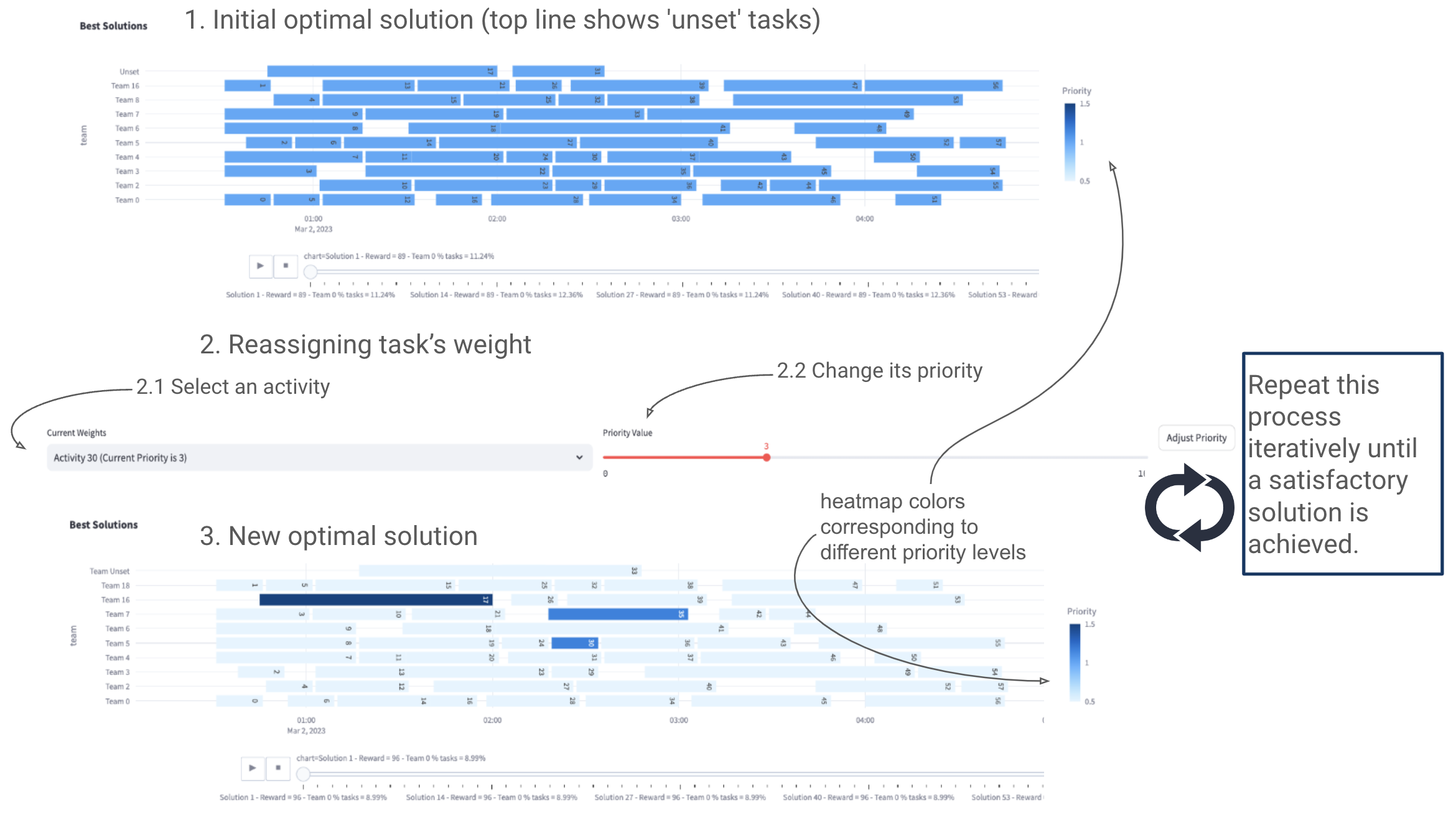}
    \caption{Conflict resolution with the fine-tuning task priorities method.
    The second image shows an example of changed priority/weight for some chosen task, leading to new solver propositions.}
    \label{conflicts_maxsat}
\end{figure}

Finally, the application provides the option to solve relaxed versions of the problem using a weighted Max-CSP formulation (optimisation variant of the satisfiability problem where each constraint is assigned a weight, and the goal is to maximize the sum of the weights of the satisfied constraints). We relax the constraint requiring each task to be allocated (constraint \ref{c-task-allocated} of our model) and maximize the sum of allocated activities, where each activity $a$ is weighted by a weight $w_a$. 
This method can easily generate several alternative solutions, maximizing the weighted objective. 
If the user is unhappy with the relaxed solutions, it is possible to interact with the solver by setting different $w_a$ weights on some chosen activities. 
This method should lead to feasible solutions obtained using domain expert constraint relaxations (see Figure \ref{conflicts_maxsat}).

\section{Conclusion \& Discussion on Future work}

Our decision-making tool for workforce allocation combines the power of constraint programming with interactive and explainable features. By involving users in the decision-making process and providing clear explanations of conflicts and resolutions, we aim to enhance the trust and adoption of CP solvers in industrial settings. The prototype application demonstrates the practical implementation of these concepts and serves as a foundation for further development and evaluation within the TUPLES project.
Our next step in the research is to evaluate the relevance of the generated explanations from a user perspective. These XAI methods should be assessed by expert users who can judge the usability and applicability of XAI/CP technology components in realistic scenarios. Hence, we plan to conduct scientifically rigorous user studies to determine preferred methods for infeasibility restoration. We also plan another user study focused more on a visual interface that will gather user feedback on conflict visualisations and description methods. We are currently implementing various visualisation approaches and textual description techniques to enhance user acceptability.

In this paper, we focused on a pure allocation problem where the activities are already scheduled and can't be shifted in time. In a more realistic model, the possibility of shifting tasks (e.g. changing the start time) in the feasibility restoration step should be considered. However, this would require to transform the model into a scheduling problem, and we are currently working in this direction. This raises interesting scalability challenges for the XAI technology bricks such as MUS computation. To address the interpretability of large conflict explanations, we could consider using step-wise explanations \cite{bleukx2023simplifying}. By breaking down complex explanations into simpler steps, we can create short, interpretable sequences that collectively clarify the issue.
Also, in this more complex setup where we consider scheduling constraints, there might be implicit constraints that the planners keep in mind but are not articulated in the problem formulation. Hence we are looking at the techniques from the literature on constraint acquisition~\cite{Tsouros_Berden_Guns_2024}. 

\bibliography{references.bib}
\end{document}